\documentclass[12pt]{article}
\usepackage{amsfonts}
\usepackage{amsmath}
\usepackage{graphicx}
\usepackage{cite}
\usepackage{url} 
\usepackage{float}
\usepackage{booktabs}
\usepackage{amsthm}

\usepackage{svg}

\theoremstyle{definition}
\newtheorem*{Example}{Example}

\newcommand{\keywords}[1]{
  \par\noindent
  \textbf{Keywords:} #1
}

\begin{document}

\title{Control of Overfitting with Physics}
\date{}

\author{Sergei V. Kozyrev\footnote{Orcid ID:0000-0002-5597-5556, Steklov Mathematical Institute of Russian Academy of Sciences, Gubkina St. 8, Moscow 119991, Russia; kozyrev@mi-ras.ru}, Ilya A. Lopatin\footnote{Orcid ID:0009-0001-8790-0540, Steklov Mathematical Institute of Russian Academy of Sciences, Gubkina St. 8, Moscow 119991, Russia; ilya.lopatin2606@gmail.com}, Alexander N. Pechen\footnote{Orcid ID: 0000-0001-8290-8300, Steklov Mathematical Institute of Russian Academy of Sciences, Gubkina St. 8, Moscow 119991, Russia; Ivannikov Institute for System Programming of the Russian Academy of Sciences, Alexandra Solzhenitsyna str. 25, Moscow, 109004, Russia; apechen@gmail.com}}

\maketitle

\begin{abstract}
    While there are many works on the applications of machine learning, not so many of them are trying to understand the theoretical justifications to explain their efficiency. In this work, overfitting control (or generalization property) in machine learning is explained using analogies from physics and biology. For stochastic gradient Langevin dynamics, we show that the Eyring formula of kinetic theory allows to control overfitting in the algorithmic stability approach---when wide minima of the risk function with low free energy correspond to low overfitting. For the generative adversarial network (GAN) model, we establish an analogy between GAN and the predator--prey model in biology. An application of this analogy allows us to explain the selection of wide likelihood maxima and overfitting reduction for GANs.
\end{abstract}

\keywords{stochastic gradient descent; generative adversarial network; overfitting control; Eyring formula; free energy; branching random process}

\section{Introduction}

Analogies from physics and other fields, particularly population genetics, are of interest when studying problems in machine learning theory. Analogies between machine learning theory and Darwinian evolution theory were discussed already by Alan Turing~\cite{Turing}. Biological analogies in computing were discussed by John von Neumann~\cite{BookNeumann}. Physical models in relation to computing were discussed by Yuri Manin~\cite{Manin}. Such analogies allow physical intuition to be used in learning theory. Among the well-known examples are genetic~\cite{BookGoldbergGA} and evolutionary algorithms~\cite{Vikhar_2016}, models of neural networks and physical systems with emergent collective computational abilities and content-addressable memory~\cite{Hopfield_1982}, a parallel search learning method based on statistical mechanics and Boltzmann machines that mimic Ising spin chains~\cite{Ackley1985}. A phenomenological model of population genetics, the Lotka--Volterra model with mutations, related to generative adversarial network (GAN) was introduced in~\cite{KozyrevGAN}. Analogies between evolution operator in physics and transformers (an artificial intelligence model) were discussed in \cite{Transformer}. Ideas of thermodynamics in application to learning were considered in \cite{LeCun, LeCun2016}  and in relation to the evolution theory in \cite{KatKoon, Vanchurin}. GANs have found many applications in physics, e.g., for designing the process of electromagnetically induced transparency metasurfaces~\cite{Zhu_Zhang_Dong_Rong_Gong_Meng_2023}, etc. Physics-informed neural networks were developed for solving the optima quantum control of open quantum systems~\cite{Norambuena_Mattheakis_Gonzalez_Coto_2022}. Beyond that, the theory of open systems inspired the development of quantum reinforcement learning, where states of the agent are quantum states of some quantum system and the dynamic environment is a quantum environment surrounding the manipulated system to optimize the reward~\cite{Dong_Chen_Tarn_Pechen_Rabitz_2008,DongTarn_2008-2}, which is connected to the general incoherent quantum control paradigm~\cite{Pechen_Rabitz_2006}. Various approaches to quantum machine learning were proposed, e.g.,~\cite{Biamonte2017}. Open quantum systems were suggested to be applied to spin glasses~\cite{Sieberer2016}, network science~\cite{Nokkala2024}, and finance~\cite{meng2016}.
Quantum neural networks were proposed to be applied for some problems, e.g., to work with encrypted data to protect the privacy of users' data and models~\cite{Sun_Chang_Wang_Zhang_Yan_2024}.

Within the framework of such analogies, it is natural to discuss biological analogs for various models of the machine learning theory. In this work, two analogies are considered, between the stochastic gradient Langevin dynamics (SGLD) in machine learning and the Eyring formula in the kinetic theory~\cite{Eyring_1935}, and between the GAN model~\cite{GAN} and the mathematical predator--prey model in biology, where we suggest to consider the discriminator and generator in GAN playing the role of prey and predator, respectively. The proposed analogies allow us to explain the efficiency of controlling overfitting, which is the lack of generalization abilities for a machine learning approach. It is known that for stochastic gradient descent (SGD), overfitting is reduced; for the GAN, this effect of reducing overfitting is even more significant. We propose to explain overfitting control in these processes within the framework of the algorithmic stability approach by suppressing narrow minima of the empirical risk function.

In Section~\ref{Sec:Stochastic}, we consider the stochastic gradient Langevin dynamics (SGLD) using a stochastic differential equation  (SDE). We show that the reduction of overfitting in such a model follows from ideas used in chemical kinetics such as the Eyring formula, which states that the reaction rate (the rate of the transition between two potential wells due to diffusion) is determined by free energy of the transition state (the saddle between two potential wells) and the free energy of the initial state of the reaction (optimization of quantities involving entropy-dependent Helmholtz free energy also appears in quantum optimization, e.g.,~\cite{Morzhin_Pechen_2023}).

In Section~\ref{Sec:GAN}, we describe the
minimax problem for the GAN model by a system of two stochastic differential equations (one is for the discriminator and another is for the generator). In this sense, the GAN model is a two-body generalization of the SGLD model considered in Section~\ref{Sec:Stochastic}. However, this generalization significantly changes the behavior of the learning system, as demonstrated by the simulations below. We show that this model implements a selection of wide maxima of the likelihood function, leading to a reduction of overfitting. A biological interpretation of the GAN model is provided in terms of the interaction between a predator and a prey (the predator is the generator, the prey is the discriminator). Learning for GANs by solving a system of ordinary differential equations was considered in \cite{Qin,Oseledets,Kolter}.

In Section~\ref{Sec:Branching}, we introduce a generalization of the GAN for a type of population genetics model. We consider a branching random process with diffusion for two types of particles (discriminators and generators), in which discriminators and generators can replicate. In this case, the rates of replication and death of particles depend on the contributions to
the functional, minimax, which is defined by the GAN model.

In Section~\ref{Sec:NR}, we provide the numerical simulations illustrating the behavior of the stochastic gradient descent and the predator--prey model of the minimax problem for the GAN for a simple potential with two wells. We observe the following two regimes: pushing out of the narrower well and oscillations in the wider well. One parameter of the interaction allows to control the transition between these two regimes. The third regime, when escape out of the both wells occurs, is possible; however, this regime can be avoided by adjusting the~parameters.

Some relevant notions from the theory of random processes are provided in \mbox{Appendix~\ref{Sec:Appendix}}. Section~\ref{Sec:Conclusions} summarizes the results.

\section{Stochastic Gradient Descent and the Eyring Formula}\label{Sec:Stochastic}

\subsection{Stochastic Gradient Descent}
Let $\{z_l\}$, $l=1,\dots,L$ (with the elements belonging to $\mathbb{R}^n$), a loss function $f_l(x)=\mathcal{L}(z_l,x)$ for the $l$-th sample object, and hypothesis $x$ (we assume that the hypothesis space is $\mathbb{R}^m$). Minimization of the empirical risk is the following problem: 
\begin{equation}\label{min}
f(x) = \frac{1}{L}\sum\limits_{l=1}^L f_l(x) \,\to\, \min\limits_{x}.
\end{equation}

Gradient descent algorithm for this problem, when the loss function belongs to the space of continuously differentiable functions $C_1(X)$, is defined by the solution of the differential equation $\frac{df}{dt}=-\frac{\partial f(x)}{\partial x}$, which is numerically described by the iterative process starting from some initial guess $x_{0}$, such that
\begin{equation}\label{gradient_descent}
x_{k+1}=x_{k} - \alpha_k f'(x_{k}) = x_{k} - \alpha_k\frac{1}{L}\sum\limits_{l=1}^{L} f'_l(x_{k}),
\end{equation}
where $\alpha_k$ is a gradient step [which is also often called as learning rate (lr)] at the $k$th iteration. There are different methods for introducing stochasticity in training for the problem~\eqref{min}, such as mini-batch learning or dropout. In this paper, we consider a method based on adding Gaussian noise to \eqref{gradient_descent}, which is related to stochastic gradient Langevin dynamics.

Stochastic gradient Langevin dynamics encompasses the modification of the above procedure where small independent random perturbations are added at each step of gradient descent, such that
\begin{equation}\label{noise_descent}
x_{k+1}=x_{k} + w_{k} - \alpha_k f'(x_{k}),
\end{equation}
where the set of all $w_{k}$ is a set of independent Gaussian random vectors.

This procedure can be considered as a discrete-time version of the stochastic equation
\begin{equation}\label{stochastic_equation}
d\xi^i(t)=\sqrt{2\theta}dw^i(t)- \frac{\partial f(\xi(t))}{\partial x^i} dt,
\end{equation}
where $dw^i(t)$ is the stochastic differential of the Wiener process (factor $\sqrt{2}$ is used for convenience to have unit instead of $1/2$ in front of $\theta\Delta$ in Equation~(\ref{diff}) below).

The stochastic gradient Langevin dynamics procedure with an equation of the form~(\ref{stochastic_equation}) was discussed in \cite{Parisi, Parisi1, Geman, Welling}.

Diffusion equation in the potential is the partial differential equation
\begin{equation}\label{diff}
{\partial u\over \partial t}=\theta\Delta u+\nabla u \cdot \nabla f+ u \Delta f,
\end{equation}
where $x\in\mathbb{R}^d$, $u=u(x,t)$ is the distribution function, $f=f(x)$ is the potential, $f\in C^2(\mathbb{R}^d)$, and $\theta>0$ is the temperature.

Equivalently, this diffusion equation can be rewritten as (where $\beta=1/\theta$ is the inverse temperature) 
\begin{equation*}
{\partial u\over \partial t}=\theta{\rm\bf div}\,\left[e^{-\beta f}\,{\rm\bf grad}\,
\left[u e^{\beta f}\right]\right].
\end{equation*}

Gibbs distribution $e^{-\beta f}$ is a stationary solution of this equation. The solution converges to the Gibbs distribution under certain conditions on $f$, as discussed in~\cite{Witten}.

Diffusion Equation (\ref{diff}) is the Fokker--Planck equation (see (\ref{Fokker--Planck}) in Appendix~\ref{Sec:Appendix}) for stochastic gradient Langevin dynamics (\ref{stochastic_equation}).

We would like to highlight the following: The well-known 
stable diffusion neural network and, in general latent, diffusion models~\cite{LatentDiffus, ChangDiffusionModel} do use diffusion described by a stochastic differential equation (as in SGLD). However, the general formulation of the problem is different there, where a special type of diffusion is used for generating objects (particularly images), while we are considering the diffusion (SGLD) for learning in a potential

\subsection{Overfitting Control for Stochastic Gradient Descent}

Overfitting is the lack of ability to generalize for the solution of the learning problem (i.e., high likelihood on the training sample and low likelihood for the validation sample). One approach to overfitting control is based on the algorithmic stability, i.e., on the stability of the solution obtained by a learning algorithm to perturbations of the training sample~\cite{Stab2,Stab3,Stab4}. In this case, narrow (sharp) minima of the empirical risk functional (in the hypothesis space) are associated with overfitting, and wide (flat) minima correspond to solutions of the learning algorithm without overfitting~\cite{Stab1}.

Introducing noise into the gradient descent procedure, i.e., considering SDE (\ref{stochastic_equation}) and diffusion (\ref{diff}), is related to the problem of overfitting as follows: The Eyring formula, which is a generalization of the Arrhenius formula, describes the reaction rate (the rate of transition between two potential wells due to diffusion of the form (\ref{diff})) in the kinetic theory: the reaction rate is proportional to
\begin{equation}\label{Eyring}
e^{-\beta(F_1-F_0)},
\end{equation}
where $F_1$ is the free energy of the transition state (the saddle between two potential wells), and $F_0$ is the free energy of the initial state of the reaction (the potential well from which the transition occurs). The free energy of a state is $F=E-\beta^{-1}S$, where $E$ is the energy and $S$ is the entropy of the state. In general, free energy of a set $U$ (say a potential well) is defined as
\begin{equation*}
e^{-\beta F(U)}=\int_{U}e^{-\beta E(x)}dx.
\end{equation*}

The connection between the Arrhenius and Eyring formulas and the spectral asymptotics for the Schr\"odinger operator corresponding to the subbarrier (tunnel) transition between two potential wells was discussed in \cite{Witten}.

Let us emphasize that we do not assume any minimization procedure---stochastic gradient Langevin dynamics (\ref{stochastic_equation}) generate the Gibbs distribution according to diffusion \mbox{Equation (\ref{diff})}, and this Gibbs distribution will be concentrated in potential wells with low free energy. In the view of the above discussion, learning with stochastic gradient Langevin dynamics is a search for a potential well that gives the global minimum of free energy. 
The influence of the temperature is important when comparing the entropy and the energy parts of free energy. For the effect of particle capture by a well (i.e., learning) to take place, it is important that the temperature should be significantly (e.g., by several times) less than the difference in the free energies of the well and the saddle; hence, the temperature is important for the SGLD. Thus, for successful learning, the temperature should be low enough to make the capture of the particle by a potential well possible.

The Eyring formula (\ref{Eyring}) implies that, for equal free energies of the transition state and equal energies of the initial state, the transition rate will be lower (the free energy of the initial state will be lower) for potential wells with higher entropy (wider ones). Thus, stochastic gradient Langevin dynamics (\ref{stochastic_equation}) will correspond to a regime in which wider wells (with higher entropy) more effectively capture the learning system, i.e., algorithmically stable solutions to the learning problem will be selected during stochastic gradient optimization (here, we consider the SGLD procedure but we believe the same effect will hold for other forms of SGD, such as mini-batch procedure).

The validity of the proposed approach is limited by the assumptions for the Eyring formula. The Eyring formula describes transitions for a diffusion equation in the potential and might be applied to various complicated high-dimensional landscapes having clear minima and saddles between them. Its validity under quite general conditions has been justified in~\cite{AvelinCMP2023}. For landscapes which might not exhibit clear minima and saddles between them, the proposed approach based on the Eyring formula may not work.

Physical analogies in machine learning, particularly the application of free energy, were discussed in \cite{LeCun}. In~\cite{LeCun2016}, the following procedure was considered: the empirical risk functional (depending on the hypothesis $x$) was replaced by the so called “local entropy” functional, which looks like minus free energy of some vicinity of $x$ (where energy is the empirical risk). In this way, the wider empirical risk minima will correspond to deeper minima of the new “local entropy” functional. Relation to the generalization property in the approach of \cite{Stab1} (flat minima) was discussed (although the authors of this paper do not discuss the Eyring formula).

\section{The GAN Model and Overfitting}\label{Sec:GAN}

\subsection{Stochastic Gradient Langevin Dynamic for GAN}
The generative adversarial network (GAN) model is a minimax problem, such that~\cite{GAN}
\begin{eqnarray}
\label{minimax}
&\min_{y} \max_{x} V(x,y);&\\
\label{Vxy}
&V(x,y)=\frac{1}{L}\sum\limits_{l=1}^L \log D(z_l,x)  + \int\limits_{Z} p_{\rm gen}(z,y) \log(1-D(z,x)) dz;&
\end{eqnarray}

$\{z_l\}$ is a sample, where $ z_l\in Z$, $D(z,x)$ and $p_{\rm gen}(z,y)$ are parametric families of probability distributions on the space $Z$, called the discriminator and the generator, with parameters $x$ and $y$ from statistical manifolds $X$ and $Y$ (which we will assume to be real vector spaces).

In \cite{GAN}, the generator was considered as a parametric family of mappings from some auxiliary space to $Z$. These mappings transferred the probability distribution on the auxiliary space to $Z$. The discriminator was described by a distribution on the same space as the data, and the interpretation was as follows: the discriminator outputs binary variable “one” or “zero” given the data and given the generator according to this distribution (i.e., outputs “one” if the discriminator considers these inputs as correct).

The first contribution to $V(x,y)$ in (\ref{Vxy}) is the $\log$-likelihood function. The second contribution behaves qualitatively as the minus inverse of the Kullback--Leibler distance (Kullback--Leibler divergence, KL distance, or KLD) between the distributions of the discriminator and the generator (that is, this contribution is negative, large in magnitude for small KL--distances, and grows to zero for large KL--distances), as was mentioned in \cite{GAN}.
\begin{equation*}
V(x,y)=V_1(x)+V_2(\rho(x,y)).
\end{equation*}
Here
\begin{equation*}
\rho(x,y)={\rm KL}(D(x)|p_{\rm gen}(y)),
\end{equation*}
where the Kullback--Leibler distance between probability distributions $p$ and $q$ is defined as
\begin{equation*}
{\rm KL}(p|q)=\int_Z p(z)\log\frac{p(z)}{q(z)}dz.
\end{equation*}

The minimax for $V(x,y)$ over $x$, $y$ is obtained from the local maximum of $V_1(x)$ over $x$. Transitions between the local maxima of $V_1(x)$ generate transitions between local minimaxes of $V(x,y)$ (the generator follows the discriminator) as shown below.

The stochastic gradient Langevin dynamics optimization for the problem  
(\ref{minimax}),
(\ref{Vxy}) can be described by a system of SDEs defining random walks $\xi=(\xi^i(t))$, $i=1,\dots,m$, $\eta=(\eta^j(t))$, $j=1,\dots,n$ on statistical manifolds $X$ and $Y$ of the discriminator and generator,
\begin{eqnarray}
\label{discriminator}
d\xi^i(t)&=&\sqrt{2\theta}dw^i(t)+ dt \frac{\partial}{\partial x^i}V(\xi(t),\eta(t)),\\
\label{generator}
d\eta^j(t)&=&\sqrt{2\theta}dv^j(t)- dt \frac{\partial}{\partial y^j}V(\xi(t),\eta(t)).
\end{eqnarray}

Here, $\frac{\partial}{\partial x^i}V$ and $\frac{\partial}{\partial y^j}V$ are derivatives of $V$ with respect to the first and second arguments, respectively; $w^i(t)$ and $v^j(t)$ are Wiener processes on the parameter spaces of the discriminator and generator. This system of stochastic equations follows directly from the minimax condition and is independent from the KL distance formulation. However, the KLD formulation is used later for analyzing the behavior of the solution. In this system of SDEs, the discriminator seeks to maximize the function $V(x,y)$ (\ref{Vxy}) with respect to $x$, and the generator seeks to minimize this function with respect to $y$.

\begin{Example}
Let us consider one-dimensional parameters $x$ and $y$ for the discriminator and for the generator, respectively, and functional $V=\omega xy$ with minimax located at the origin. The noiseless GAN equation system is
\begin{eqnarray*}
\frac{dx}{dt}&=&\frac{\partial}{\partial x}V(x,y)=\omega y,\\
\frac{dy}{dt}&=&-  \frac{\partial}{\partial y}V(x,y)=-\omega x.
\end{eqnarray*}

Its solution is
\begin{equation*}
x=A\sin \omega (t-t_0),\quad  y= A \cos\omega (t-t_0)
\end{equation*}
with oscillations around the minimax.
\end{Example}

In~\cite{Qin,Oseledets,Kolter}, the convergence of the optimization of the GAN model  by the gradient descent method with respect to the parameters of the discriminator and generator in the neighborhood of the functional's local minimax was studied, and oscillations of the parameters were discussed.

\subsection{Overfitting Control for GAN}
If we ignore the presence of the generator, then the dynamics of the discriminator (\ref{discriminator}) for optimization with noise will correspond to the diffusion in the potential generated by the data. Thus, the arguments of Section \ref{Sec:Stochastic} will be applicable. Therefore, overfitting can be reduced according to the Eyring formula.

The presence of the generator will further suppress overfitting. The minimax problem for the GAN (\ref{minimax}) can be described as follows. The discriminator (\ref{discriminator}) tries to reach regions of the parameter $x$ with high values of $V(x,y)$. The generator (\ref{generator}) tries to reach regions of the parameter  $y$ with low values of $V(x,y)$. In this case, the contribution to $V(x,y)$ (\ref{Vxy}) from the likelihood function depends only on the parameters of the discriminator, i.e., the discriminator tries to increase both contributions to (\ref{Vxy}), and the generator tries to decrease only the second contribution. The second contribution to (\ref{Vxy}) decreases at small Kullback--Leibler distances between the discriminator and the generator.

Therefore, the compromise between the optimization problems for the discriminator and the generator will be achieved when they are located at maxima of the contribution from the likelihood function to (\ref{Vxy}) which are sufficiently wide in the space of the parameters $(x,y)$ (where the average KL--distance between the discriminator and the generator is not too small). Selecting wide maxima, in accordance with the algorithmic stability approach, will reduce the effect of overfitting.

Here, we propose a biological prey--predator interpretation for the GAN model, which is completely different from the interpretation used in \cite{GAN}. In our interpretation, the discriminator is herbivore (prey), the generator is predator, and the data are grass. Then, the minimax problem (\ref{minimax}) describes the situation when the discriminator searches for grass (the maximum of the likelihood function; this corresponds to an increase in the first contribution in (\ref{Vxy})) and also runs away from the predator (this corresponds to an increase in the second contribution to (\ref{Vxy})), while the predator chases the prey (and hence decreases this contribution). In our interpretation, the discriminator (herbivore)---as a distribution---tries to get closer to the data distribution (grass) and farther from the generator (predator) as a distribution  (in the KL distance sense), and the generator tries to be closer to the discriminator. As a result of this interaction, the generator also moves toward the data (grass) because herbivores (discriminator) are likely to be found there; however, this does not mean that the predator tries to imitate grass (this is a mixture of the two interpretations of the GAN). Minimization in (\ref{minimax}) for the generator forces the predator to move to fields (or meadows, likelihood maxima) where the discriminator is present. The interaction of the two contributions to (\ref{Vxy}) forces the discriminator to search for sufficiently wide meadows (likelihood maxima) where the average KL--distance from the predator is not too small. In general, the predator pushes out the prey from narrow fields of grass, and both the prey and predator move to wide grass fields. Thus, the GAN model implements the selection of wide likelihood maxima, which reduces overfitting.

Simulations illustrating the discussed above behavior are considered in Section \ref{Sec:NR} below.

\section{Branching Random Process for GAN}\label{Sec:Branching}

In this section, a branching random process with diffusion and particle interactions describing the populations of discriminators and generators in a generalization of the GAN model is introduced.

The theory of branching random processes and its connection with population genetics have been actively discussed in the literature, for example, in~\cite{Sevast,Vatutin}. Previously, in \cite{KozyrevGAN}, a generalization of the GAN model related to population genetics (a Lotka--Volterra-type model with mutations), was discussed. In this model, discriminators and generators could reproduce and form populations. The phenomenological equations of population dynamics were considered, and the suppression of overfitting was discussed.

Consider a generalization of the GAN to the case of several discriminators (particles in the hypothesis space of the discriminator with parameter $x$, particles are indexed by $a$) and generators (particles in the hypothesis space of the generator with parameter $y$, particles are indexed by $b$).   
The analog of the SDE system (\ref{discriminator}), (\ref{generator}) will take the form
\begin{eqnarray}
\label{discriminator1}
d\xi^{i}_{(a)}(t)&=&\sqrt{2\theta}dw^i_{(a)}(t)+ dt \frac{\partial}{\partial x^i}V(\xi_{(a)}(t),\overline{\eta(t)});\\
\label{generator1}
d\eta^j_{(b)}(t)&=&\sqrt{2\theta}dv^j_{(b)}(t)- dt \frac{\partial}{\partial y^j}W(\overline{\xi(t)},\eta_{(b)}(t));
\end{eqnarray}
where each particle is associated with its own independent Wiener process $w^i_{(a)}(t)$, $v^j_{(b)}(t)$ on the right-hand side of the equation in the discriminator and generator spaces, respectively, and the terms with interaction on the right-hand sides of the equations have the~form
\vspace{-24pt}

\begin{eqnarray}\label{V1}
V(x,\overline{y})&=&\frac{1}{L}\sum_{l=1}^L \log D(z_l,x)  + \sum_{b} \int_{Z} p_{\rm gen}(z,y_b) \log(1-D(z,x)) dz = V_1(x,\{z\})+V_2(x,\overline{y});\qquad\\
\label{W1}
W(\overline{x},y)&=&\sum_{a}\int_{Z} p_{\rm gen}(z,y) \log(1-D(z,x_a)) dz.
\end{eqnarray}
Here, $V_1(x,\{z\})$ is the likelihood function for the discriminator $x$.

This corresponds to a GAN-type model with functional $V(x,\overline{y})$ for discriminator $x$ and functional $W(\overline{x},y)$ for generator $y$. Equations (\ref{discriminator1}) and (\ref{generator1}) describe optimization by the stochastic gradient Langevin dynamics. Each discriminator interacts with a set of generators and similarly, each generator interacts with a set of discriminators. Here, $\overline{x}=\{x_1,\dots,x_M\}$ and $\overline{y}=\{y_1,\dots,y_N\}$ are sets of discriminators and generators, respectively. The second contribution $V_2(x,\overline{y})$ and the function $W(\overline{x},y)$ contain sums from contributions that behave qualitatively as $-{\rm KL}(x,y)^{-1}$, where ${\rm KL}(x,y)$ is the Kullback--Leibler distance between discriminators and generators with parameters $x$ and $y$.

Let us define a model which mimics the population genetics, defined by a branching random process with diffusion and interaction with particles of two types $\xi^{i}_{(a)}(t)$, $\eta^j_{(b)}(t)$ (discriminators and generators), which can perform random walks in accordance with Equations (\ref{discriminator1}) and (\ref{generator1}), and have the ability to replicate and die, with the probabilities of such processes depending on the functionals (\ref{V1}), (\ref{W1}). The replication of a particle consists of replacing it with two particles of the same type with the same coordinates (which can then perform random walks in accordance with (\ref{discriminator1}), (\ref{generator1})).

We propose to use the following branching rates (as related to Lotka--Volterra-type model discussed in \cite{KozyrevGAN}): the death rate of generators is considered as fixed, while the replication rates of the generators $\eta_{(b)}(t)$ are proportional to (recall that both $W$ and $V_2$ are~negative)
\begin{equation*}
-W(\overline{\xi(t)},\eta_{(b)}(t));
\end{equation*}

the replication rates of discriminators $\xi^{i}_{(a)}(t)$ are proportional to
\begin{equation*}
\exp\left(V_1(\xi^{i}_{(a)}(t),\{z\})\right);
\end{equation*}
the rate of death of the discriminator $\xi^{i}_{(m)}(t)$ is proportional to
\begin{equation*}
-V_2(\xi^{i}_{(a)}(t),\overline{y}).
\end{equation*}

Thus, discriminators replicate depending on the data and die depending on the generators. Generators replicate depending on the discriminators and die at a constant~rate.

The biological interpretation of the proposed model is the following. The data $\{z_l\}$ are the distribution of grass, the discriminators $\xi^{i}_{(a)}(t)$ are herbivores (prey), the generators $\eta^j_{(b)}(t)$ are predators; herbivores reproduce on grass, and predators hunt herbivores. The effect of suppressing overfitting on narrow likelihood maxima looks as follows: if the discriminator has replicated on the likelihood maximum (on its statistical manifold $X$), the generator will tend to go there and replicate there (the generator will tend to the corresponding regions of its statistical manifold $Y$, such that the KL--distance between $D(\cdot,x)$ and $p_{\rm gen}(\cdot,y)$ is small). In this case, for a narrow likelihood maximum, the average KL--distance will be small, i.e., the predator will eat the prey more effectively (and then suffer from hunger) than for a wide maximum. This is how the effect of selective suppression of narrow population maxima in $X$ and $Y$ corresponding to narrow likelihood maxima is realized. For the case of the population genetics model for the GAN (where discriminators and generators can replicate), the effect of overfitting control is more pronounced than for the standard GAN model (without replication).

\section{Simulations}\label{Sec:NR}
In this section, the results of the numerical simulation of the SGLD procedure and the simulation of the predator--prey model for the GAN are provided.

\subsection{Objective Function}

Let $\mathcal L(x)$ be an objective function for optimization, ${\mathcal L}(x) \to \max$. We consider ${\mathcal L}(x)$ as a sum of non-normalized Gaussians of the following form:
\begin{equation} \label{eq:NumericalSimulationTargetFunc}
\mathcal L (x) = \sum_{j=1}^n q_j e^{ - \frac{ \| x - c_j \|^2 }{2 \sigma_j^2} },
\end{equation}
where $q_j$ and $\sigma_j$ are some real-value positive constants, $x, c_j \in \mathbb{R}^d$. For visualization, we use $d=2$. Constant $\sigma_j$ can be interpreted as a characteristic width of the extremum. We consider the case when $ \sigma_k \ll \| c_i - c_j \| $ for all $k, i\ne j$, so that ${\mathcal L}(x)$ has $n$ separated extrema.

This objective function has two hills  around the two extrema (or wells for the minimization of $-{\mathcal L}(x)$; for definiteness, we call them below as wells).

\subsection{Stochastic Gradient Langevin Dynamics}

\begin{figure}[h]
\centering
\includegraphics[width =1\linewidth]{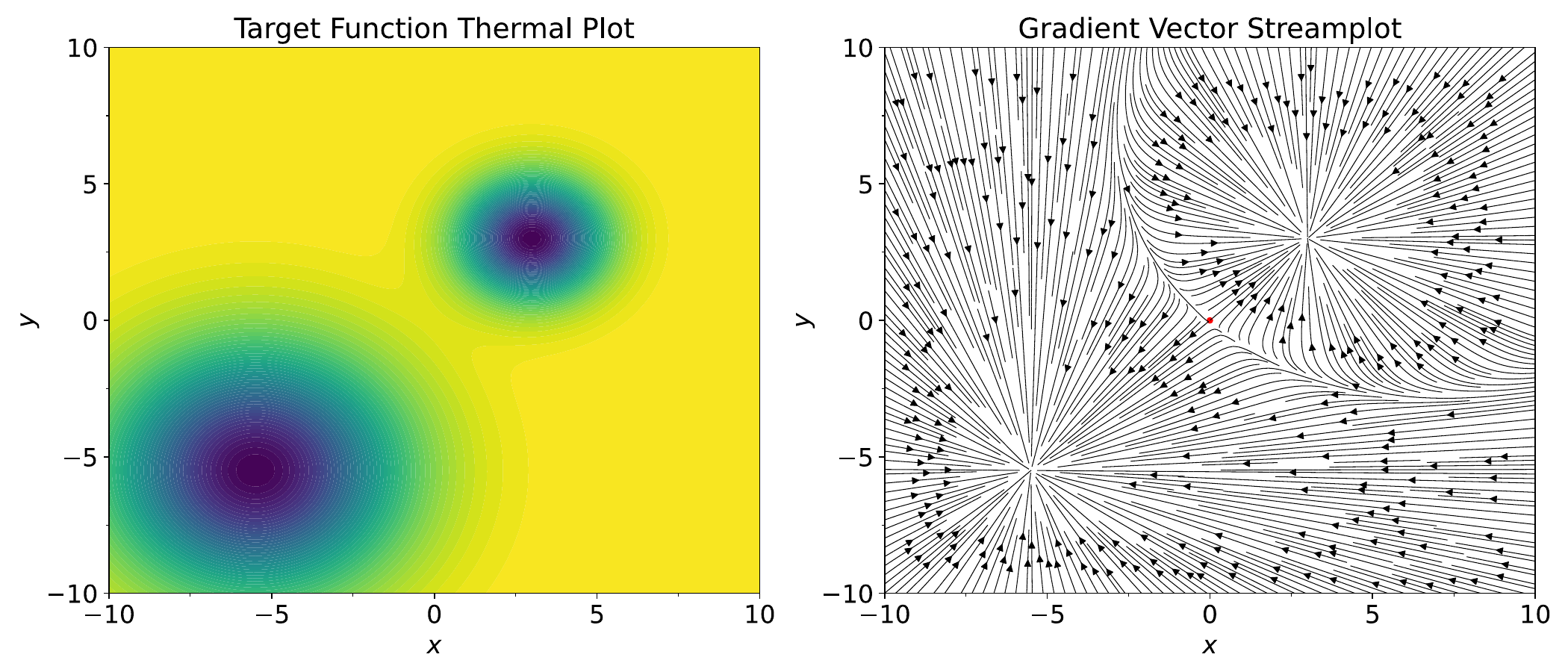}
\caption{Thermal plot of the function $\mathcal L$ (\textbf{left}) and its gradient field (\textbf{right}). Red dot in the center of the gradient field plot shows the starting point $(0,0)$.}
\label{Fig:GraphTargetFunc}
\end{figure}

For objective function~\eqref{eq:NumericalSimulationTargetFunc}, we consider $n=2$, $\sigma_1 = 3.0, \sigma_2 = 1.5$, $c_1 = \left(-5.5, -5.5 \right)^{  \mathop{T} },\linebreak  c_2 = \left(3.0, 3.0 \right)^{  \mathop{T} }$, $q_j = \sigma_j^2$ for $j = 1, 2$ (here and in similar places below, $T$ means transpose of the vector, not temperature). Thermal plot of the function $\mathcal L$ and its gradient field are shown on Figure~\ref{Fig:GraphTargetFunc}. As the starting point, set $\left( 0.0 , 0.0 \right)^{  \mathop{T} }$. Note that $\nabla_x {\mathcal L}( 0, 0)$ has positive coordinates. Therefore, standard gradient descent procedure starting from the point $x_0=\left( 0.0 , 0.0 \right)^{  \mathop{T} }$ will converge to $c_2$, which is the well with a smaller width. Consider standard stochastic  gradient~procedure
\begin{equation*}
x_{k+1} = x_k  + \alpha \nabla_x {\mathcal L}( x_k) + \xi_k, \quad x_0 = (0, 0)^{\mathop{T}},
\end{equation*}
where $\xi_k$ are two-dimensional independent random variables,
\begin{equation*}
\xi_k \sim ( {\cal N} (0, T (1+k)^{-1/2} ),{\cal N} (0, T (1+k)^{-1/2} )),
\end{equation*}
where ${\cal N}$ is the normal distribution with center at the origin and with variance $T (1+k)^{-1/2}$, $T$ is temperature parameter, and $\alpha$ is a fixed learning rate. The scaling $(1+k)^{-1/2}$ is introduced to ensure convergence of the SGLD process.
\vspace{-4pt}

\begin{figure}[h]
\centering
\includegraphics[width = .6\linewidth]{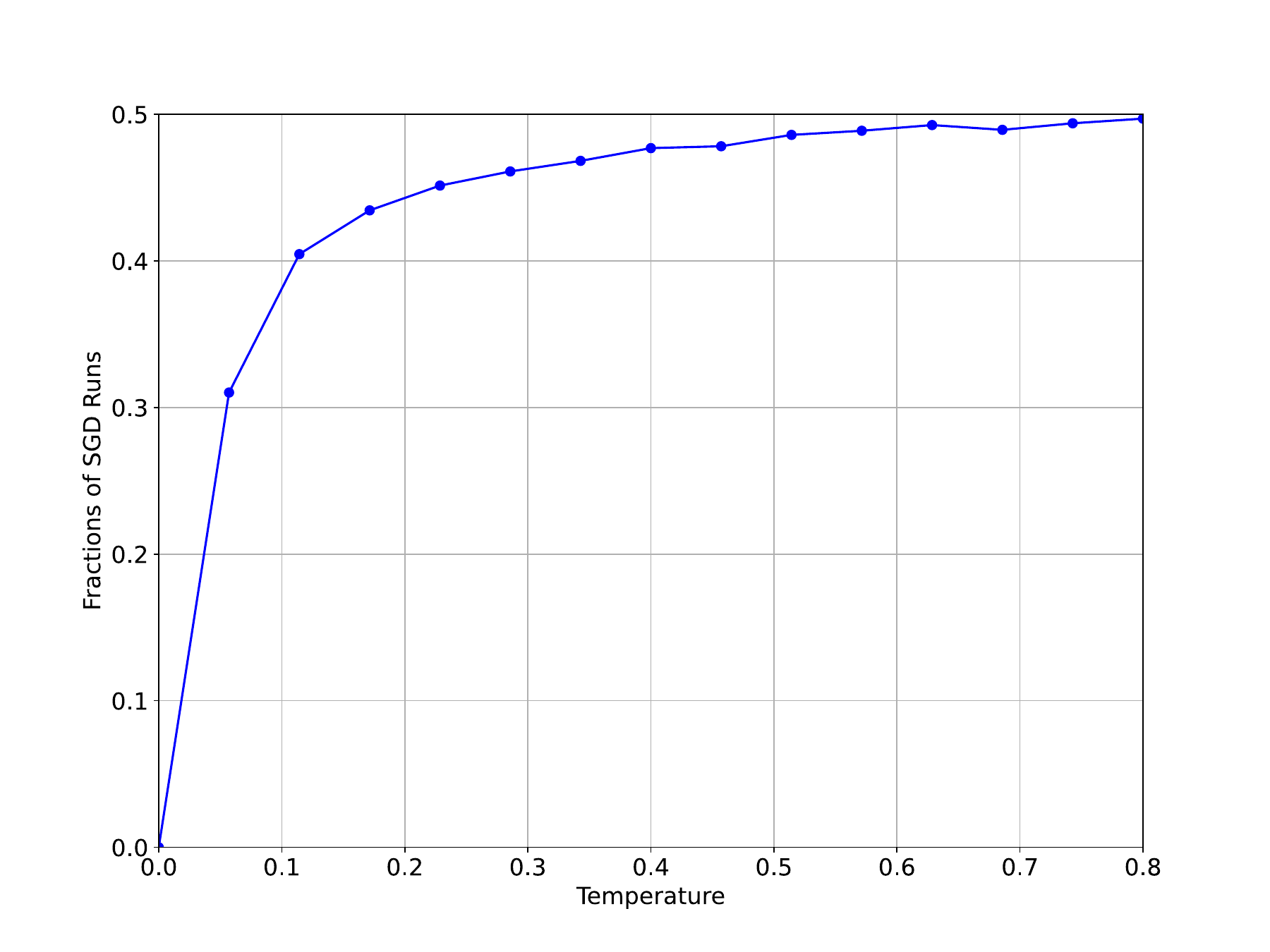}
\caption{Fraction of the runs of the SGLD starting at the point $x_0$, which converge to the well $c_1$ with greater width.
}
\label{fig:SGD-simulation}
\end{figure}

We consider 15 values of temperature $T$ uniformly spanned on $[0, 0.8]$. For each value of $T$, we run the SGLD  starting from the same initial point $x_0$  with a maximal number of iterations $K=2000$ and select those runs which in no more than $K$ iterations converge to one or another well (in their $\sigma$-vicinities). Then, we compute the fraction of the runs which converge to the wider well. For $T=0$, all runs converge to the well with a smaller width. According to the Eyring formula \eqref{Eyring}, we expect an increase in the fraction of the SGLD's runs from the starting point $x_0$ with increasing temperature $T$, which converge to the well $c_1$ with a larger width. Figure~\ref{fig:SGD-simulation} confirms this behavior.

\begin{figure}[h]
\centering
    \includegraphics[width= .8\linewidth]{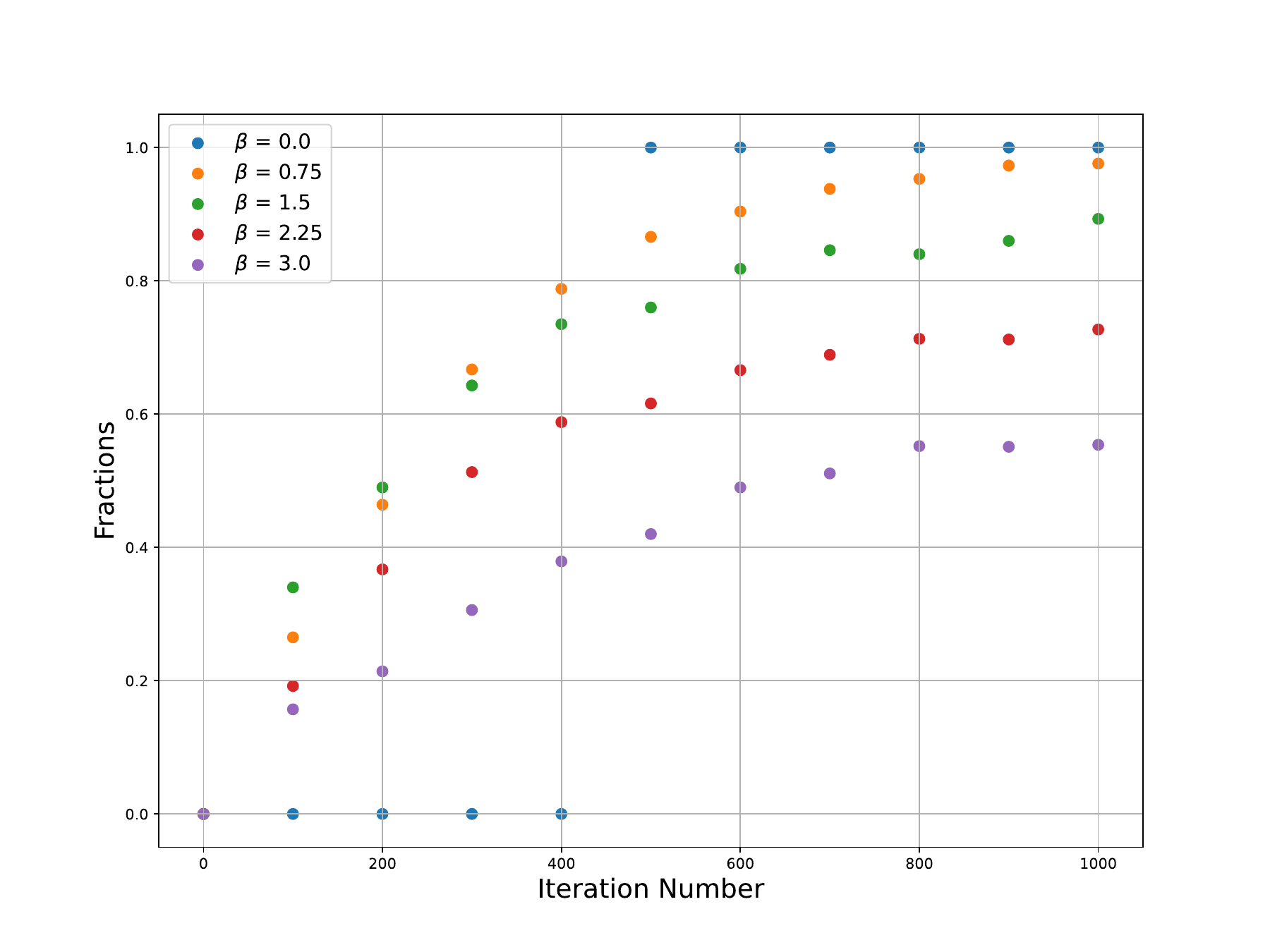}
    \caption{Fraction of the points which converge to extrema vs. iteration number plotted for several inverse temperatures $\beta=0, 0.75, 1.5, 2.25, 3.0$.}
    \label{fig:RatioPoints}
\end{figure}

The efficiency of SGLD depends on such hyperparameters as temperature and learning rate, which influence both the convergence behavior and the stability of the solution SGLD. Since  our analysis is limited to the use of a stochastic differential equation, we just use a sufficiently low learning rate value without a detailed analysis and tuning of its value.
The conditions on the learning rate are that it should be sufficiently small to guarantee not escaping the target well and also it should be not very small, since if it is very small, it will require a large number of iterations.
Temperature is relevant for our consideration. An increase in the temperature affects the convergence and stability of the SGLD because if the temperature is too high, fluctuations can throw the trajectory beyond the minimum. To demonstrate this, we plot the dependence of the trajectories' fractions on Figure~\ref{fig:RatioPoints} that converge to one of the two extrema on the number of iterations for different temperatures and for a single learning rate.
\vspace{-3pt}

\vspace{-10pt}

\subsection{Predator--Prey Model}

For the predator--prey model, we consider a more general dynamical system than the system defined by Equations~\eqref{discriminator} and \eqref{generator}. Let $x(t)$ be position of the prey and $y(t)$ be position of the predator at time $t$. In simulations for visualization, we consider $x,y\in\mathbb R^2$. We consider their joint evolution as governed by the system of equations
\begin{equation}
   \left\{ \begin{aligned}
        \frac{d x}{dt} &= \mathcal{L} (x) + V ( x(t) - y(t) ), \\
        \frac{d y}{dt} &= - W ( x(t) - y(t) ),
    \end{aligned} \right.
\end{equation}
where $V$ and $W$ are some vector functions (forces) describing the interaction between the prey and the predator (in this subsection, they are not the same as functions $V$ and $W$ considered in the previous sections). Informally speaking, $x(t)$ tries to evolve in a way to simultaneously maximize the objective function $\mathcal{L}$ and the distance to the predator, while $y(t)$ tries to evolve in a way to minimize the distance to the prey. For $W$, we choose
\begin{eqnarray}
    W(x-y) = \alpha_y \frac{x-y}{ \| x-y \|},
\end{eqnarray}

which defines the motion of $y$ with a constant speed $\alpha_y$ towards $x$.

A key point in our analysis is to find suitable conditions on the vector function $V(x(t) - y(t))$, which depends on the difference $x(t) - y(t)$. Let $d = \| x(t) - y(t) \|$ be the distance  between $x(t)$ and $y(t)$. We suggest the potentials to have the following general behavior at various distances:
\begin{itemize}
    \item  At short distances, $d  < \sigma_{\rm min}$, where $\sigma_{\rm min}$ is an estimate of the minimum acceptable width of the well (i.e., some width from which we do not consider the well to be narrow). For this distance, we assume $\| V \| \gg 1 $ to allow a predator to push the prey out of the well.
    \item At intermediate distances, $ \sigma_{\rm min} < d < \sigma_{\rm max} $, where $\sigma_{\rm max}$ is the estimate from above of the width of the well of $\mathcal{L}(x)$. For these distances, we assume $\| V\| \sim \| W \|$. This condition is introduced to have oscilations in sufficiently wide wells.
    \item  At long distances, $d >  \sigma_{\rm max}$, we assume $V \sim 0$ to guarantee convergence to some well of $\mathcal{L}(x)$.
\end{itemize}

\begin{figure}[h]
\centering
\includegraphics[width = 0.79\linewidth]{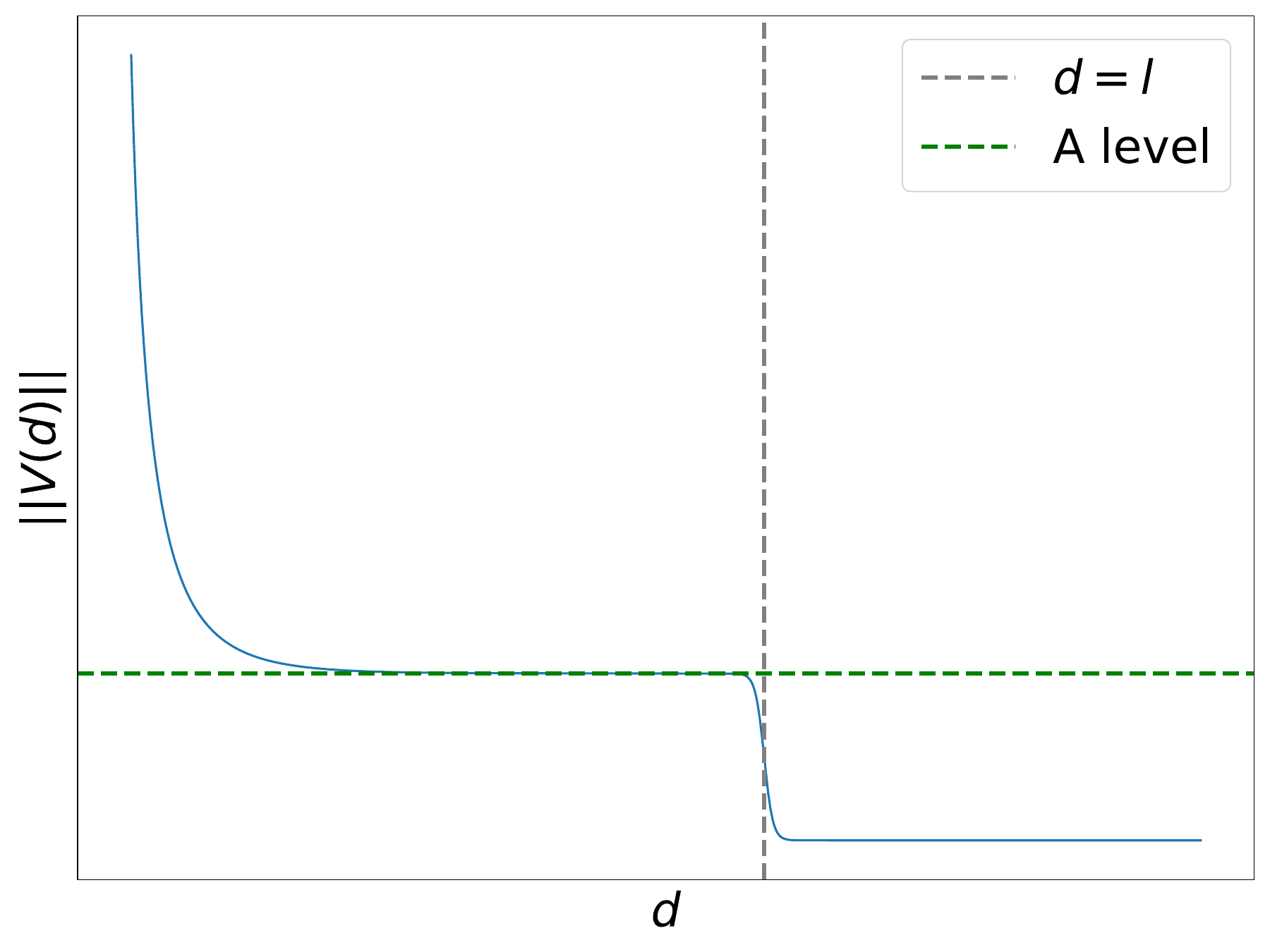}
\caption{Absolute value of the norm of the vector function $ \| V(d) \|$ and its characteristic points: $l$ and $A$.}
\label{Fig:Potential}
\end{figure}

As an explicit potential which satisfies these conditions, we take
\begin{equation} \label{eq:Vx-GeneralForm}
V ( x - y) = \left( \frac{A}{1 + e^{c (d-l) } } + C  \frac{ e^{-\sigma d} }{d}  \right) \frac{x-y}{d},\quad d = \|x-y\|.
\end{equation}

with some parameters $A, C, \sigma, c, l>0$. It contains the Yukawa potential $\frac{ e^{-\sigma d} }{d} $ as a summand. The parameter $l$ can be interpreted as a characteristic intermediate distance and $A$ as a mid-range predator, as shown in Figure~\ref{Fig:Potential}. The norm of the interaction vector function $V$ along with its characteristic points for the parameter values described below is plotted in Figure~\ref{Fig:Potential}.

\begin{figure}[h]
\centering
    \includegraphics[width=0.5\linewidth]{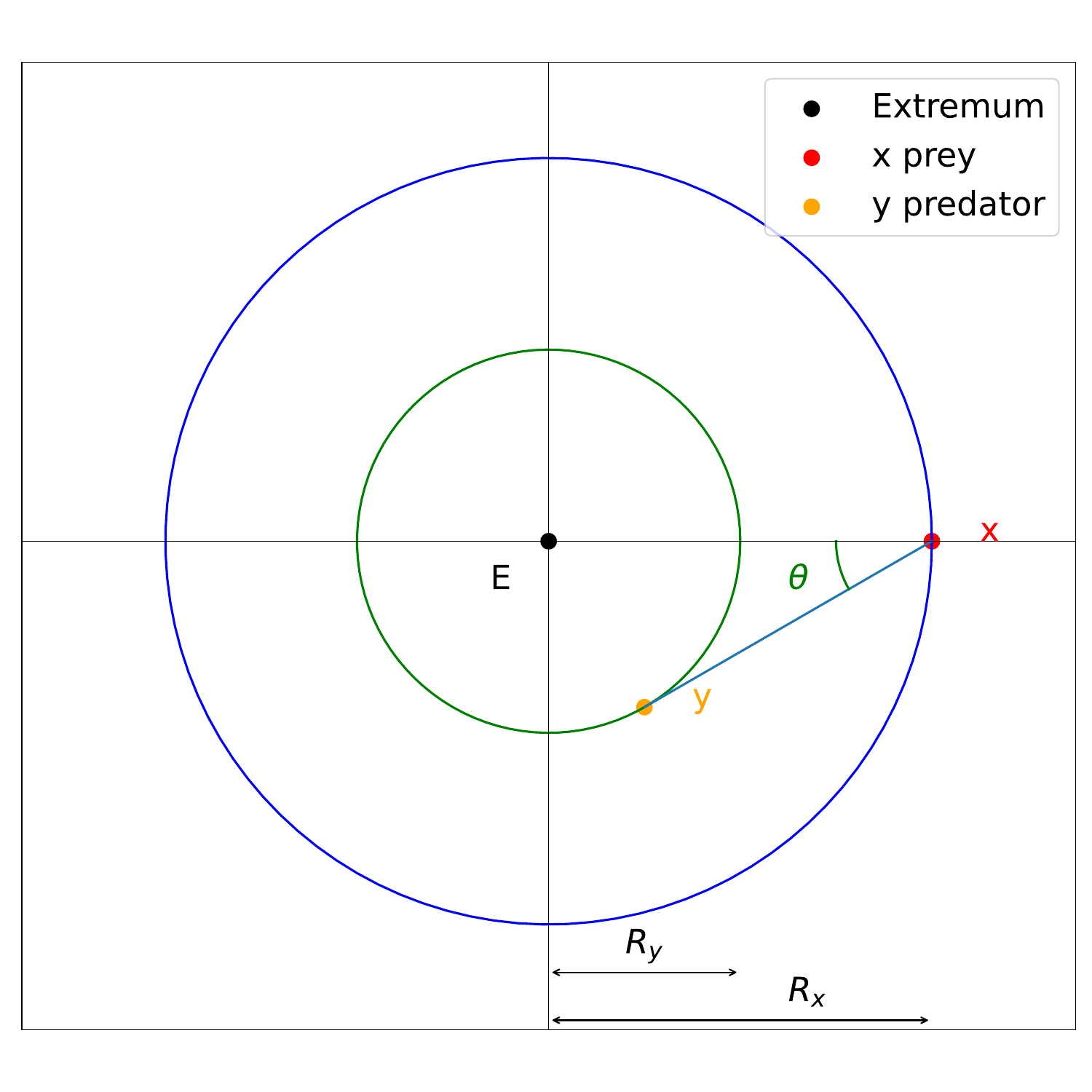}
    \caption{The limiting unstable oscillations around the extremum point for the predator--prey model.}
    \label{fig:LimitingCyrclesPredPreyMod}
\end{figure}

We assume that by tuning the parameters of the interaction vector functions $V$,  
we can push the system out of narrow wells of ${\mathcal L}$. To estimate these parameters, we consider the case of limiting oscillations around a radially symmetric well. Schematically, such oscillations can be represented in Figure \ref{fig:LimitingCyrclesPredPreyMod}. Note that this behavior is not stable for small values of $\theta$ due to the discreteness of the step also known as learning rate. Let $R_x$ and $R_y$ be distances between the extremum and $x$ or $y$, respectively, and let $\theta$ be the angle between the predator--prey line and the radial line. In our simulations, the smallest, critical angle value is about $ \theta \approx \pi /18$. For this behavior with limiting oscillations and radially symmetric structure, we have the following relation:
\begin{equation} \label{eq:LimitOcillation}
\left\{ \begin{aligned}
    & \frac{R_x}{R_y} = \frac{ \| \nabla_x {\mathcal L} (x)  + V (x, y) )  \|_{\|x\| = R_x, \|x-y\| = d }  }{\alpha_y}, \\
    & \| \nabla_x {\mathcal L} (x) \|_{\|x\| = R_x} = \|   V (x, y) \|_{ \|x-y\| = d}, \\
    & R_y^2 = R_x^2 + d^2 - 2 R_x d \cos(\theta).
\end{aligned} \right.
\end{equation}

To enable pushing out of narrow wells and oscillations inside sufficiently wide wells, the constants $\alpha_y, A, l, c, C, \sigma$ should satisfy the following heuristic conditions:
\begin{itemize}
    \item  The constant $A$ should be small enough not to generate a too strong pushing-out potential; otherwise, the prey would escape all the wells.
    \item The constant $A$ should satisfy $A > \alpha_y$, so that $x$ can keep at some distance from $y$.
    \item The constant $l$ should correspond to a sufficient width of the well. If $l$ is too large, the non-convergence to any well can occur because $x$ will enter the regime of running away from $y$. If $l$ is too small, the dynamics will not have oscillations in the wider~well.
    \item The constant $\sigma$ in~Equation~\eqref{eq:Vx-GeneralForm} determines the minimal width of a well.
    The width of the well from which there will be pushout due to the short-range Yukawa potential is defined as $\sim$$\sigma^{-1}$ (up to some constant).
    \item The constant $C$ must be chosen to be large enough so that the Yukawa potential creates a repulsion stronger than the attraction of the gradient $\mathcal{L}$ near the narrow~well.
    \item The parameter $c$ characterizes the rate of the transition from intermediate to long distances; its value is taken to be large enough. With increasing $c$, the norm $\|V(d)\|$ of the vector function $V(d)$ tends to be more step-like with a gap at $d=l$.
\end{itemize}

\vspace{-10pt}

A typical evolution is shown in Figure~\ref{fig:GANsimulation} (video available in Supplementary Material ). The model parameters for this simulation are the following. Centers of the two Gaussian extrema in~\eqref{eq:NumericalSimulationTargetFunc} ($n=2$) are $c_1 = (0, 0), c_2 = (-7.0, -7.0)$, widths $\sigma_1 = 0.5, \sigma_2 = 2.0$, and amplitudes $q_1 = 0.25, q_2 = 8.0$. The parameters of the function $V$ defined by Equation~\eqref{eq:Vx-GeneralForm} are the following: $A = 0.3$, $l = 1.0$, $ c = 10^3$, $C = 10.0$, $\sigma = 10.0$. Starting points for the prey and for the predator are $(0.5, 0)$, $(0, 2.0)$, respectively.

We observe the possibility of having the following two regimes: pushing out of the narrower well and oscillations in a wider well. One of the parameters of the interaction vector functions controls the transition between these two regimes. It has the meaning of maximal width of the well from which pushout is expected. The third regime, when escape out of the both wells occurs, is possible, but we were able to overcome this regime by adjusting the parameters.

\begin{figure}[h]

\includegraphics[width=.8\linewidth]{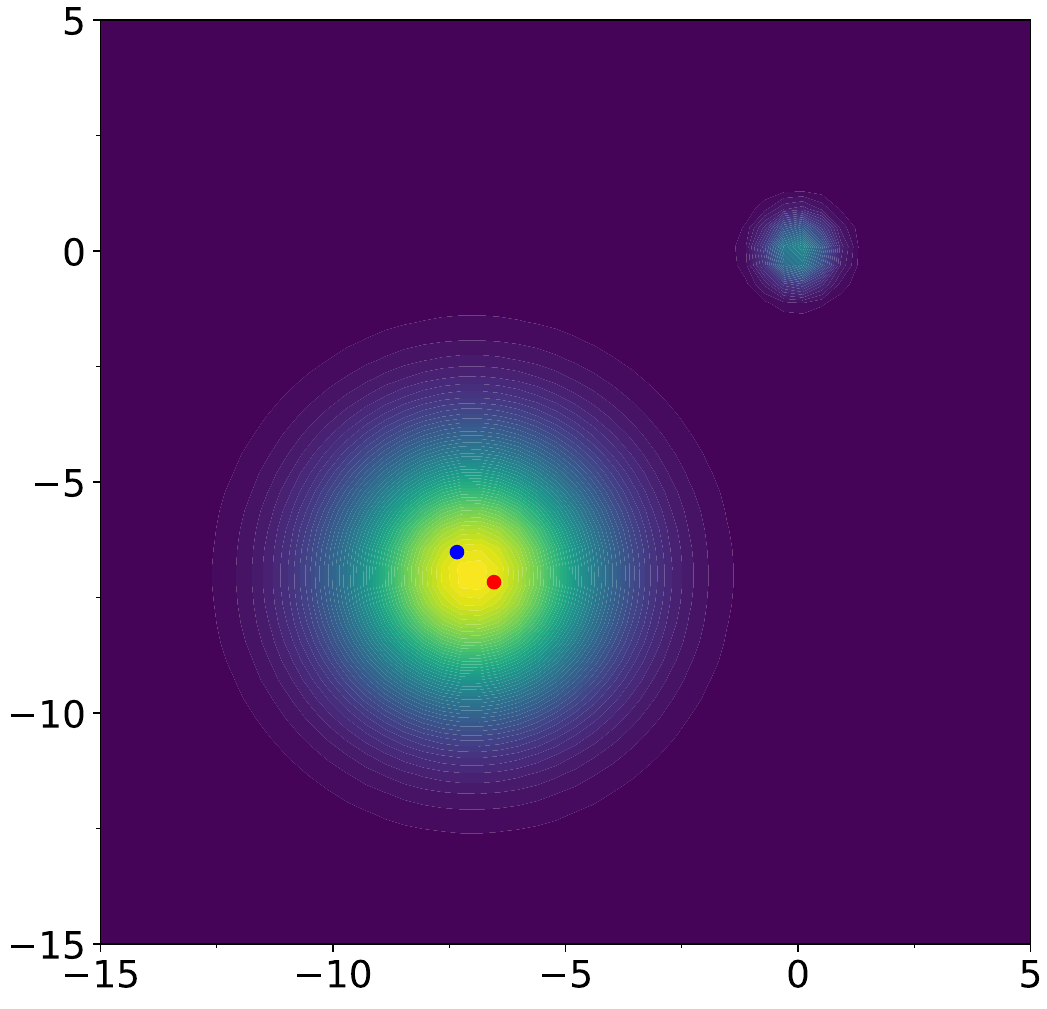}
\centering
\caption{Simulation  of the GAN process for the potentials $V$ and $W$ with two extrema of ${\mathcal L}(x)$ having different widths (video available in Supplementary Material ). The red dot shows the prey and the blue dot the predator. We observe two regimes: first, the prey escapes of the narrow well and moves to the wider well (first regime), where it starts to oscillate (second regime).}
\label{fig:GANsimulation}
\end{figure}

\subsection{Application to Wine Recognition Dataset} \label{sec:ToyDataset}

In Section \ref{Sec:NR}, we applied the predator--prey model to some synthetic conditions to reveal the desired behavior. While the application of the method to real big datasets is a separate complex task, here we investigate the improvement achieved by the suggested method on an educational small dataset. As an example, we consider the Wine recognition dataset from scikit-learn library (\url{https://scikit-learn.org/stable/datasets/toy_dataset.html\#wine-recognition-dataset},accessed on 10.10.2024). This dataset consists on 178 instances with 13 numeric attributes. Target is class attribute which is encoded by numbers $0, 1, 2$. To make the overfitting process more visible and for simplicity of having overfitting, we consider this dataset as a regression task. Linear regression is a good solution for this dataset; however, to create an overfitting situation, we consider quadratic regression with ordinary gradient descent and compare it with the predator--prey model. We randomly split the dataset on train and test sets (train is $80\%$) and run gradient descent for linear and quadratic regression and predator--prey model for quadratic regression. The obtained results, summarized in Table~\ref{tab:ToyDataSet}, show that the predator--prey model allows to improve the learning results. Running with other random splittings of the dataset into train and test subsets, which show similar results in general. However, several differences may occur. First, linear regression sometimes shows better results than quadratic regression. Second, at first, iteration gradient descent sometimes occurs faster that the predator--prey model convergence to some minimum, but then GD goes up and starts overfitting, while the predator--prey model oscillates around the minimum that avoids overfitting. Thus, in general, our finding on this example is that the predator--prey model works better and sometimes much better to reduce the effect of~overfitting.

\begin{table}[h]
\caption{Results of the work of linear and quadratic regression with gradient descent, as well as the predator--prey model on the Wine recognition dataset. MSE loss is mean squared error; accuracy is ratio of the corrected predicted class in percent.}
\label{tab:ToyDataSet}
\centering
\begin{tabular}{lcc}
\toprule
            & \textbf{MSE Loss} & \textbf{Accuracy} \\
\midrule
linear GD & 0.12 & 81\% \\
quadratic GD & 0.12 & 89\% \\
quadratic PP-model & 0.07 & 94\% \\
\bottomrule
\end{tabular}
\end{table}

\section{Conclusions}\label{Sec:Conclusions}

Various mimics of physical or biological behavior do appear in machine learning, e.g., in evolutionary and genetic algorithms. In this work, we discuss a possible justification, based on some models appearing in physics and biology, for the ability to control overfitting in SGLD and the GAN. For SGLD, we show that the Eyring formula of the kinetic theory allows to control overfitting in the algorithmic stability approach, when wide minima of the risk function with low free energy correspond to low overfitting. We also establish a relation between the GAN and the predator--prey model in biology, which allows us to explain the selection of wide likelihood maxima and overfitting reduction for the GAN (the predator pushes the prey out of narrow likelihood maxima). We performed numerical simulations and suggested conditions on the potentials which would imply such behavior.

\section{Appendix A} \label{Sec:Appendix}
Here, we provide some relevant notions from the theory of random processes~\cite{Ventzel}.

{\bf Fokker--Planck equation}. 
Consider a diffusion with a generator
\begin{equation}\label{gen}
\hat L f(x)=\frac{1}{2}\sum_{ij}a^{ij}(x)\frac{\partial^2 f(x)}{\partial x^i \partial x^j}+\sum_{i}b^i(x)\frac{\partial f(x)}{\partial x^i},
\end{equation}
and the adjoint operator
\begin{equation*}
\hat L^* f(x)=\frac{1}{2}\sum_{ij}\frac{\partial^2 \left(a^{ij}(x) f(x)\right)}{\partial x^i \partial x^j}-\sum_{i}\frac{\partial \left( b^i(x) f(x)\right)}{\partial x^i}.
\end{equation*}

Then, the transition diffusion probability density satisfies the Fokker--Planck equation

\begin{equation}\label{Fokker--Planck}
\frac{\partial p(t,x,y)}{\partial t}=\hat L^*_y p=\frac{1}{2}\sum_{ij}\frac{\partial^2 \left(a^{ij}(y) p(t,x,y)\right)}{\partial y^i \partial y^j}-\sum_{i}\frac{\partial \left( b^i(y) p(t,x,y)\right)}{\partial y^i}.
\end{equation}

A stochastic differential equation
\begin{equation}\label{stochastic}
d\xi^{i}(t)=\sum_{j}\sigma^i_j(\xi(t))dw^j(t) + b^i(\xi(t))dt,
\end{equation}
where $dw^j(t)$ is the stochastic differential of the Wiener process, defines diffusion with a generator of the form (\ref{gen}), where
\begin{equation*}
a^{ij}(x)=\sum_k \sigma^i_k(x)\sigma^j_k(x),
\end{equation*}
and $b^i$ in ((\ref{gen}), (\ref{Fokker--Planck})) coincide with $b^i$ in (\ref{stochastic}).

\end{document}